\documentclass[10pt,twocolumn,letterpaper]{article}

\usepackage[pagenumbers]{cvpr} 

\usepackage{graphicx}
\usepackage{amsmath}
\usepackage{amssymb}
\usepackage{booktabs}

\usepackage[pagebackref,breaklinks,colorlinks]{hyperref}

\usepackage[capitalize]{cleveref}
\crefname{section}{Sec.}{Secs.}
\Crefname{section}{Section}{Sections}
\Crefname{table}{Table}{Tables}
\crefname{table}{Tab.}{Tabs.}

\def\cvprPaperID{*****} 
\def\confName{CVPR}
\def\confYear{2023}

\begin{document}

\title{Self-Supervised Representation Learning for CAD}

\author{%
  Benjamin T. Jones\thanks{http://www.bentodjones.com} \\
  University of Washington\\
  Seattle, WA 98195 \\
  \tt\small{benjones@cs.washington.edu} \\
\and
  Michael Hu \\
  University of Washington\\
  Seattle, WA 98195 \\
  \tt\small{mkhu@cs.washington.edu} \\
\and
  Vladimir G. Kim\thanks{http://www.vovakim.com} \\
  Adobe Research\\
  Seattle, WA 98103 \\
  \tt\small{vokim@adobe.com} \\
\and
  Adriana Schulz\thanks{https://homes.cs.washington.edu/$\scriptstyle\sim$adriana} \\
  University of Washington\\
  Seattle, WA 98195 \\
  \tt\small{adriana@cs.washington.edu} \\
}
\maketitle

\begin{abstract}
    The design of man-made objects is dominated by computer aided design (CAD) tools. Assisting design with data-driven machine learning methods is hampered by lack of labeled data in CAD's native format; the parametric boundary representation (B-Rep). Several data sets of mechanical parts in B-Rep format have recently been released for machine learning research. However, large scale databases are largely unlabeled, and labeled datasets are small. Additionally, task specific label sets are rare, and costly to annotate. This work proposes to leverage unlabeled CAD geometry on supervised learning tasks. We learn a novel, hybrid implicit/explicit surface representation for B-Rep geometry, and show that this pre-training significantly improves few-shot learning performance and also achieves state-of-the-art performance on serveral existing B-Rep benchmarks.
\end{abstract}

\begin{figure*}
  \includegraphics[width=\textwidth]{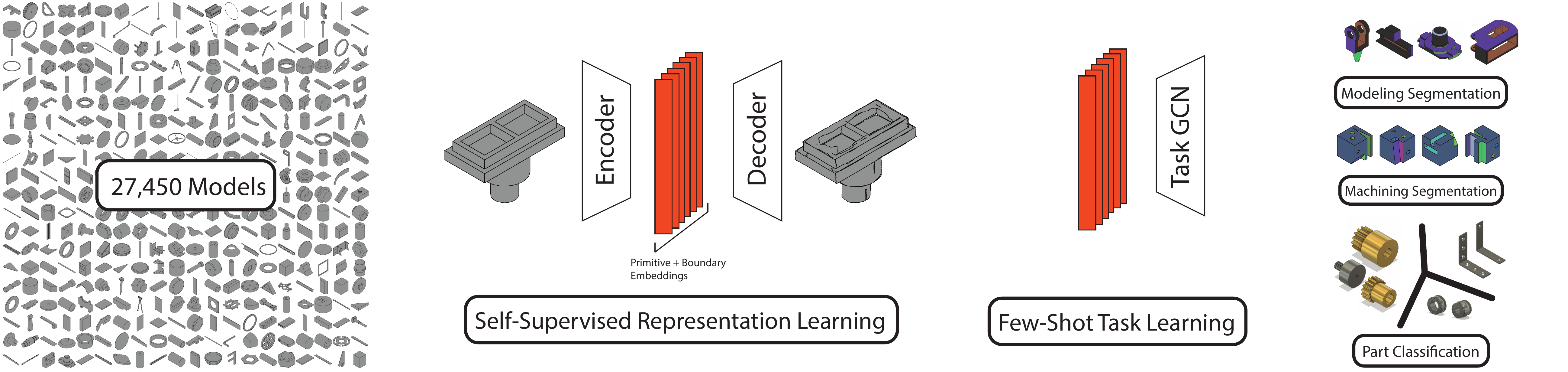}
  \caption{Overview of our technique. We train a geometric self-supervision task of a large, unlabeled dataset of CAD Boundary Representations (B-Rep) to learn geometrically relevant representations for each B-Rep face. These pre-trained representations are then used to train few-shot segmentation and classification learning tasks on labeled B-Rep datasets.}
  \label{fig:overview}
\end{figure*}

\section{Introduction}
 
Almost every human-made object that exists today started its life as a model in a CAD system. As the prevalent method of creating 3D shapes, repositories of CAD models are extensive. Further, CAD models have a robust structure, including geometric and program representations that have the potential to expose design and manufacturing intent. Learning from CAD data can therefore enable a variety of applications in design automation and design- and fabrication-aware shape reconstruction and reverse engineering. An important challenge in learning from CAD is that most of this data does not have labels that can be leveraged for inference tasks. Manually labeling B-Rep data is time consuming and expensive, and the specialized format requires CAD expertise, making it impractical for large collections. 

In this work we ask: how can we leverage large databases of \emph{unlabeled} CAD geometry in analysis and modeling tasks that typically require labels for learning? 

Our work is driven by a simple, yet fundamental observation: the CAD data format was not  developed to enable easy visualizing or straightforward geometric interpretation, it is a format designed to be compact, have infinite resolution, and allow easy editing. Indeed CAD interfaces consistently run sophisticated algorithms to convert the CAD representation into geometric formats for rendering. Driven by this observation our key insight is to leverage large collections of unlabeled CAD data to learn to geometrically interpret the CAD data format. We then leverage the networks trained over the geometric interpretation task in supervised learning tasks where only small labeled collections are available. In other words, we use geometry as a model of \emph{self-supervision} and apply it to \emph{few-shot-learning.}

\begin{figure}[h!]
	\centering
	\includegraphics[width=\linewidth]{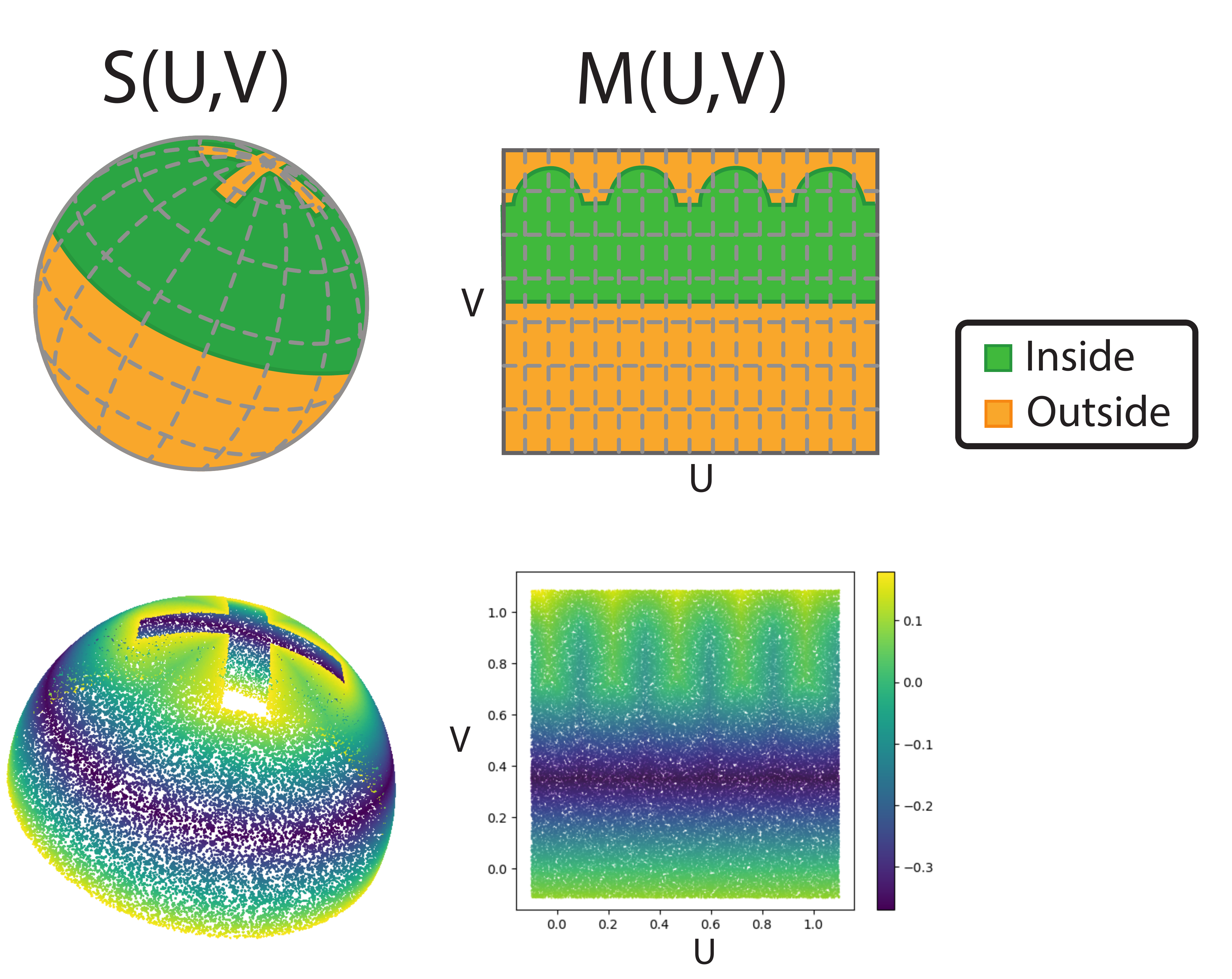}
	\caption{Face representation in B-Rep geometry. Above: schematic of a parameterized surface (sphere, left) with its unrolled clipping plane (right). Adjacent edges (dark green) implicitly define the  clipping boundary. Below: sampled signed distance field (SDF) of the clipping plane in flattened (right) and embedded (left) form; this is the function we actually learn.}
	\label{fig:brepclip}
\end{figure}

Specifically, we learn to rasterize local CAD geometry using an encoder-decoder structure.  The standard CAD format encodes geometry as parametric Boundary Representations (B-Rep). B-reps are graphs where the nodes are parametric geometry (surfaces, curves, and points) and edges denote the topological adjacency relationships between the geometry. Importantly, the parametric geometry associated with each node is \emph{unbounded} and bounds are computed from the topological relationships: curves bounding surfaces and points bounding curves. As shown in Figure~\ref{fig:brepclip}, the geometry of B-rep face is computed by \emph{clipping} the surface primitive, where the clipping masked is constructed from adjacent edges.

This means that  B-reps are constructed pieccewise by \textit{explicitly} defined surfaces with \textit{implicitly} defined boundaries. This is the observation that drives our proposed learning architecture: reconstructing faces by jointly decoding the explicit surface parameterization as well as the implicit surface boundary. Our proposed encoder uses message passing on the topological graph to capture the boundary information to encode B-Rep faces. To handle the graph heterogeneity (nodes comprised of faces, edges, and, vertices), we use a hierarchical message passing architecture inspired by the Structured B-Rep GCN~\cite{jones:2021:automate}. Our decoder uses the learned embeddings as latent codes for two per-face neural function evaluators: one mapping from $\mathbb{R}^2\to\mathbb{R}^3$ that encodes the face's parametric surface, and one mapping $\mathbb{R}^2\to\mathbb{R}$ that encodes the face's boundary as a signed distance field (SDF) \textit{within} the parametric surface. 

We apply our proposed model of B-rep self-supervision to learn specialized B-Rep tasks from very small sets of labeled data---10s to 100s of examples, vs 10k to 100k. This is done by using the embeddings learned on self-supervision as input features to supervised tasks. We evaluate our approach on three tasks and data-sets from prior work~\cite{cao2020graph,lambourne_brepnet_2021,fabwave2019}. We validate our findings across varying training set sizes and show that our model consistently outperforms the prior supervised approaches with a significant improvement in performance on smaller training sets. This result shows that self-supervision on unlabeled CAD data can enable supervised learning on small labeled data-sets paving the way to many exciting application is the domain---from modeling, to analysis, reconstruction and beyond. By using less data, our approach is also significantly faster to train, enabling applications that depend on training speed. Since our self-supervision task in a differentiable CAD rasterizer, we use it to prototype a reverse engineering task, demonstrating how it can be used in the future for different applications that require gradient-based optimization over B-Reps.

\section{Related Work}

\paragraph{Learning from CAD Collections}
Recently, interest in learning from CAD data has increased due to advances in machine learning and the release of large CAD data-sets, including collections of B-reps and program structure~\cite{koch_abc_2019,willis_fusion_2020}, CAD sketches~\cite{seff_sketchgraphs_2020}, and CAD assemblies~\cite{jones:2021:automate}. Prior work has leveraged such collections for many applications including segmentation~\cite{cao2020graph}, classification~\cite{fabwave2019}, assembly suggestions~\cite{willis2021joinable,jones:2021:automate}, and generative design of CAD sketches~\cite{seff_sketchgraphs_2020,ganin2021computer,para2021sketchgen,seff2021vitruvion}, B-Reps~\cite{jayaraman2022solidgen,Guo2022Complexgen}, and CAD programs~\cite{wu2021deepcad,willis_fusion_2020}. There is, however, a fundamental gap between the capabilities shown in past work and real-world applications that can fundamentally impact CAD design, such as auto-complete modeling interfaces or reconstruction of complex geometries: the lack for task-specific labels in large datasets. For example, while most prior work leveraged the Onshape Public dataset\cite{koch_abc_2019}, this dataset contains mostly designs created by novice CAD users, not capturing the design process of CAD experts. The Fusion 360 dataset~\cite{willis_fusion_2020} is significantly smaller compared to the Onshape collection, and other public resources, such as GrabCAD~\cite{grabcad}, contain designs from multiple CAD systems that are mostly unlabeled. In this work, we make a fundamental step toward enabling novel applications on CAD learning by proposing a novel direction for leveraging unlabeled data in supervised learning of CAD geometry.

\paragraph{Learning on B-Reps.}
A CAD B-Rep is a specialized data structure that encodes solid geometry as a graph of parametric shapes and their topological relationships, which has the advantages of arbitrary spatial resolution and easy programmatic editability. B-reps can be exported to other common geometric representations, such as polygonal meshes, using geometric CAD kernels~\cite{parasolidCAD}. Several techniques for learning on B-reps use message passing networks. BRepNet~\cite{lambourne_brepnet_2021} and and UV-Net~\cite{jayaraman2021uv} create a reduced graph of B-Rep faces, while SB-GCN~\cite{jones:2021:automate} propose a hierarchical structure over the 4 classes of topological entities (faces, loops, edges, and vertices). These methods, however, require a CAD kernel in the loop to generate the features from the B-Rep format and large labeled sets for learning. While we use a CAD kernel to train our encoder-decoder on unlabeled data, our task-specific networks can be trained on small sets of labeled data without the need of a CAD kernel at inference time.

\paragraph{Neural Shape Representations.}
Neural shape generation is an active research area with a large number of representations used by prior techniques. Some methods operate over a fixed discretization of the domain into points~\cite{achlioptas2017latent_pc,fan2017}, voxel grids~\cite{DBLP:journals/corr/BrockLRW16,liu2018voxelgan}, or vertex coordinates of a mesh template~\cite{liu2018meshVAE}. Due to irregularity of geometric data, functional representation is often used, learning to represent shapes as continuous functions, such as surface atlases~\cite{groueix2018atlasnet,yang2018foldingnet} or signed distance fields defined over a volume~\cite{park2019deepsdf,mescheder2019occupancy,chen2019implicitfields}. In this work we chose to use functional representation of the output shape as a neural occupancy and 3D mapping function over UV domains. Functional representation is well-suited for heterogeneous geometry with varied levels of detail, since it does not require choosing a fixed sampling rate. Furthermore, our representation that combines implicit fields and atlas-like embeddings directly produces a surface and can be easily supervised with the ground truth B-Rep data.

\paragraph{Few Shot Learning.}
Performance of strongly-supervised methods is commonly hampered by lack of training data. Few-shot learning techniques often rely on learning rich features in a self-supervised fashion, and then using a few examples to adapt these features to a new task~\cite{yaqing_fewshot}. One common self-supervision strategy is to withhold some data from the original input, and train a network to predict it. For example, one can remove color from images and train a network to colorize~\cite{larsson2016learning,zhang2016colorful}, remove part of the image and train a network to complete it~\cite{pathak2016context}, randomly perturb orientation and predict the upright position~\cite{gidaris2018unsupervised}, or randomly shuffle patches and predict their true ordering~\cite{doersch2015unsupervised,noroozi2016unsupervised}. Another commonly used tool is auto-encoders that encode input to a lower-dimensional space and then attempt to reconstruct it with a decoder~\cite{kingma2014semi,kingma2013auto}. Our approach is closest to the auto-encoders, except our input and output are in two different representations: CAD B-Reps and surface rasterizations. This enables us to learn features related to the actual 3D geometry.

\section{Geometric Self-Supervision}\label{sec:selfsup}

Our goal is to learn relevant features of CAD B-Reps that can be used on a number of different modeling and analysis tasks. We use an encoder-decoder architecture on B-Rep surfaces to learn a latent space of relevant features. Based on the insight that the learned features should include geometric understanding of CAD shapes, we train our encoder-decoder to learn to rasterize CAD models. We further choose to learn this embedding at the face level as opposed to a feature per part. This is driven by our application domain, where tasks typically require understanding local topological information. An overview of or architecture is shown in Figure~\ref{fig:overview}.

\paragraph{Decoder}
As previously discussed, the format we selected for the decoder output is driven by the observation that B-Reps are compositions of explicitly represented surfaces with implicitly represented boundaries. For example, the geometry associated with faces are unbounded parametric surfaces $S:\mathbb{R}^2 \to \mathbb{R}^3$ and the boundary of such surfaces is captured by the geometry of the neighboring edges. Edges in turn are defined by unbounded parametric curves bounded by neighboring vertices. Vertices are represented as points. In addition to the domain parameters, parametric geometry has additional, fixed parameters (like radius or cone angle) that we call \emph{shape parameters}. A bounded surface is called a \textit{clipped surface}, and the bounding function, which is defined \textit{implicitly} by the bounding topology, is its \textit{clipping mask}, illustrated in Figure~\ref{fig:brepclip}.

For self-supervision, we need to choose an output geometric representation for surface patches with boundary. Bounded patches do not have a natural parameterization and so are not suitable to learn directly as parametric functions. They also are difficult to represent as neural implicits.~\cite{palmer2022deepcurrents} However, the clipping function itself defines a closed region \textit{within} the parameterization of the supporting surface, which is a function that lends itself well to implicit representation in 2D. We therefore choose to learn an implicit function of the clipping region as a 2D signed distance function. Since this representation relies on an explicit surface parameterization, we also learn the supporting surface parameterization. Crucially, this does not require parameterizing a boundary.

We choose to use a conditional neural field as our decoder, since this representation can capture both explicit and implicit geometry. Specifically, the explicit parametric surface is an $\mathbb{R}^2\to\mathbb{R}^3$ function, mapping  $(u,v)$ coordinates of a face's supporting surface to the 3D position of that point $(x,y,z)$. The clipping mask encoded as an SDF over the parametric domain is a function $\mathbb{R}^2\to\mathbb{R}$, that maps the same $(u,v)$ coordinates to a value  $(d)$ measuring the signed distance to the boundary. We combine these two function to learn a single field that maps $\mathbb{R}^2\to\mathbb{R}^4$. We parameterize this field as a 4-layer fully-connected ReLU network where the input coordinates and conditioning vector are concatenated to the output at every hidden layer (similar to DeepSDF~\cite{park2019deepsdf}). The conditioning vector is the output of our face encoder described below. We call the evaluation of this field ``rasterization'' because raster sampling the field and filtering by $d$ yields a 3D surface rasterization.

To normalize the field input range, and constrain the uv-space we must sample while rasterizing, we reparameterize the uv-space of each surface prior to training so that the clipping mask fits snugly within the unit square ($[0,1]^2$). This way, our encoder-decoder is learning the explicit surface, the implicit boundary, and the support range of the implicit boundary mask.

\paragraph{Encoder}
Our encoder design is driven by the same observation that B-Reps are compositions of explicitly represented surfaces with implicitly represented boundaries. For the encoder, this means that we must capture adjacent topological entities. We propose to encode this using message passing on the topological graph. 

\begin{figure}
	\centering
	\includegraphics[width=\linewidth]{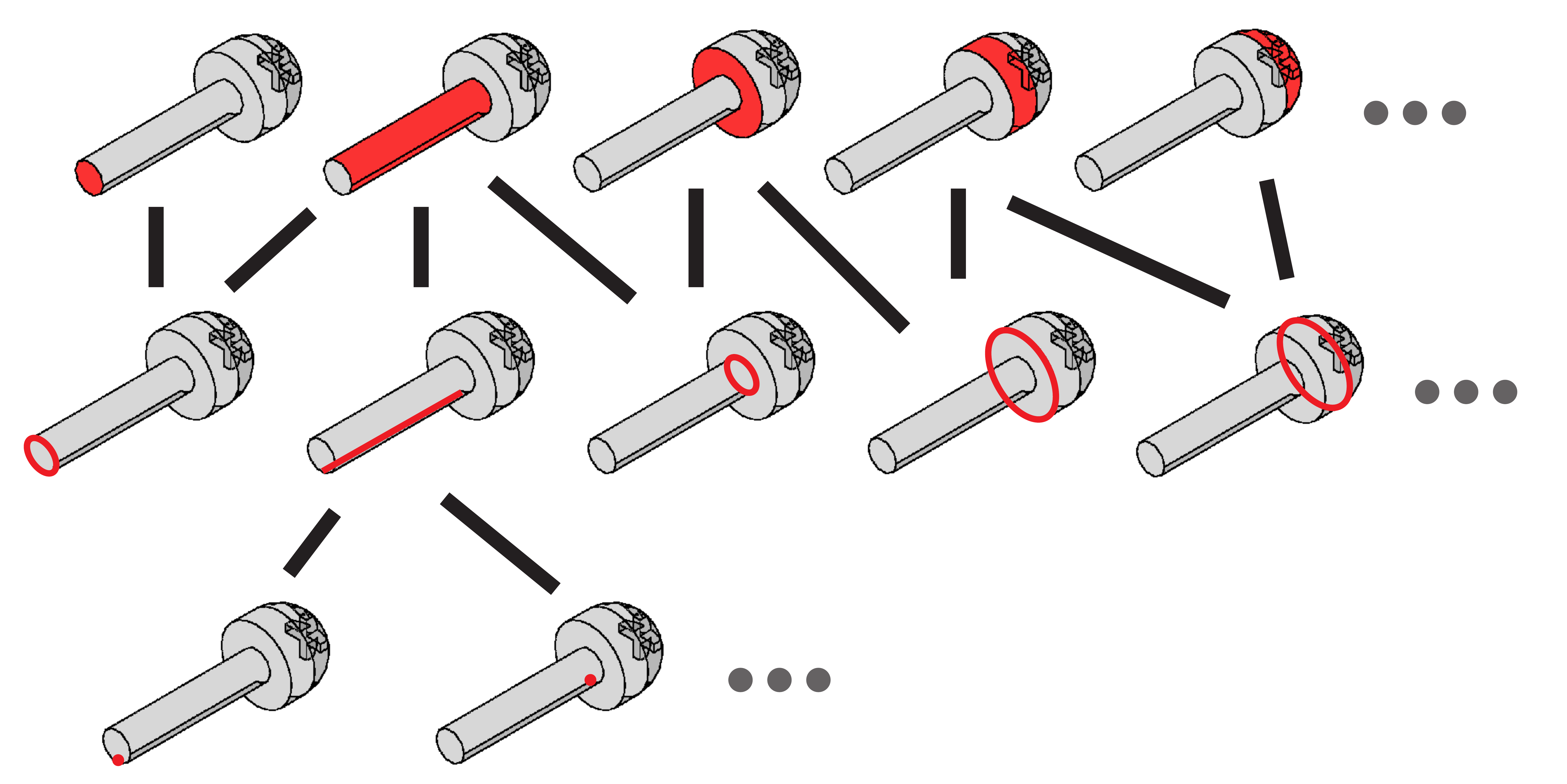}
	\caption{A boundary representation is made of parametric surface patches, bounded by parametric edges, bounded by vertices. The relationship between these entities forms a hierarchical graph.}
	\label{fig:brepdecomp}
\end{figure}

As show in Figure~\ref{fig:brepdecomp}, the B-rep topological graph has a hierarchical structure; high dimensional topological elements are only adjacent to the immediate lower dimensional entities: faces are adjacent to their bounding edges, which are adjacent to their end-point vertices. Driven by this observation, use as our face-encoder a hierarchical message passing network inspired by the Structured B-Rep GCN (SB-GCN)~\cite{jones:2021:automate}: a graph convolutional network that leverages the hierarchical structure of B-reps to learn embeddings for each topological entity.

From SB-GCN, we adopt the idea of passing messages up from bounding entities to those they bound (vertices to edges and edges to faces). Since we are only attempting to learn local features, we stop message passing at this stage. Unlike SB-GCN, we only use 3 layers of topological entities -- vertices, edges, and faces -- avoiding an intermediate \emph{loop} level that aggregates edges into closed curves labeled by whether they are interior or exterior face boundaries. To achieve a similar goal in our architecture we use multi-headed graph attention to handle large and variable numbers of B-Rep graph neighbors, and graph-edge features indicating the type of relationship between B-Rep topological entities (is a face to the left or right on a B-Rep edge, and is a vertex the start or end of a B-Rep edge).~\cite{you2020designspace}

Finally, SB-GCN (and other B-Rep representation learning networks) use evaluated geometry information as input features in addition to the pure parametric geometry, such as surface bounding box and area. These features are computed by a CAD kernel: modeling software that understands how to construct and evaluate CAD geometry. Since we want to force our network to learn the evaluation function of a CAD kernel, as well as be fully differentiable back to the shape parameters, we only use the parametric geometry definitions as input features. We encode each entity with attached geometry (vertices, edges, and faces) as a one-hot encoding of the type of surface or curve concatenated with the continuous parameters of the function (for example, radius, center point, and axis for a cylinder), and a flag indicating orientation of the entity relative to the geometry (is the curve parameterization or surface normal reversed). Geometric functions with fewer parameters are zero-padded to create a fixed size input vector for each topological entity. In order to have a fixed input size, we limit our input to B-Reps that have geometry with a fixed number of parameters: planes, cylinders, cones, spheres, and tori for surfaces, and lines, circles, and ellipses for edges. We validated this choice by filtering the parts in the Automate data set~\cite{jones:2021:automate}, and found that 72.16\% contained only these primitives.

\paragraph{Training}
We trained our encoder-decoder to minimize $L^2$ losses at randomly sampled uv-points on each face's supporting surface. We used a biased random sampling in the re-parameterized uv-square $(u,v) \in [-0.1,1.1]^2$ to ensure positive and negative SDF samples were collected. CAD kernels support efficiently querying if a point in a surface parameterization is inside or outside the clipping function, so we approximate the SDF by sampling a large number of uv-points (5000) and matching each inside and outside point to its nearest neighbor in the opposite set, using a KD-Tree to accelerate these queries. We keep $N = 500$ of these points on each face for training. To bias our sample towards the 0-level set, as is common when training implicit SDFs~\cite{park2019deepsdf}, we sort by $|d|$ and task $40\%$ of our sample to be the nearest points to the boundary, and randomly sample the rest.

We trained our geometric self-supervision over 23266 shapes from the training set of Fusion 360 Gallery Segmentation Dataset~\cite{willis_fusion_2020}. Optimization was performed using the Adam optimizer~\cite{kingma2014adam} with a learning rate of $.0005$. For our experiments, we used an embedding size of 64 for all message passing layers, 2 attention heads for the vertex to edge layer, 16 attention heads for the edge to face layer, and a hidden size of 1024 for all self-supervision decoder layers.

\section{Few-Shot Learning}\label{sec:fewshot}

Since in typical task-specific applications it is challenging to find or construct large collections of labeled CAD data, we propose levering our rich latent space to enable supervised learning over very small collections. In addition to enabling training on very little data our approach also ensures that we do not need to use computationally expensive CAD kernel functionality for generating the features at inference time.

We propose two few-shot learning setups. The first are B-Rep segmentation tasks, which assign a task-specific label to each face of a B-Rep; for example, what type of machining technique can be used to construct that surface. The second are B-Rep classification tasks, which assign a label to an entire B-Rep; for example the mechanical function of that part (gear, bracket, etc.) We frame both of these as a multi-class classification problems.

\paragraph{B-Rep Segmentation Network}
Since our encodings are learned at the face level, they can be used directly as input for face level predictions. Because classification of a face is often dependent on its context within a part, we use a small graph convolutional network to capture this context; a 2-layer Residual MR-GCN.~\cite{li2020sketch2cad} We use pre-computed face embeddings from our geometric self-supervised learning as node features, and use face-face adjacency as edges; we construct this graph by removing vertex nodes from the B-Rep graph and contracting edge nodes, preserving multiple edges between faces if they exist. Output predictions for each face are then made with a fully connected network with two hidden layers.

\paragraph{B-Rep Classification Network}
While we do not have latent codes for entire B-Rep shapes, we take a similar approach to previous B-Rep learning architectures and pool learned features for each face. To do this we project each face's embedding into a new vector for pooling using a fully connected layer, followed by a second projection of the pooled part features into the prediction output space.

\section{Results}

We validate the application of our proposed approach by applying our method to three few-shot learning tasks. We further evaluate our method by analyzing the rasterization results.   

\subsection{Few-Shot Construction-Based Segmentation}
The first task we apply our method to is segmentation of B-Rep geometry by the modeling operation used to construct each face (extrusion, revolution, chamfer, etc.). This requires the model to understand how geometry is constructed in CAD software. For this task we use 27450 parts from the the Fusion 360 Gallery dataset, a collection of user-constructed parts annotated with one of 8 face construction operations on each face~\cite{lambourne_brepnet_2021}.

\begin{figure}
    \centering
    \includegraphics[width=\linewidth,keepaspectratio]{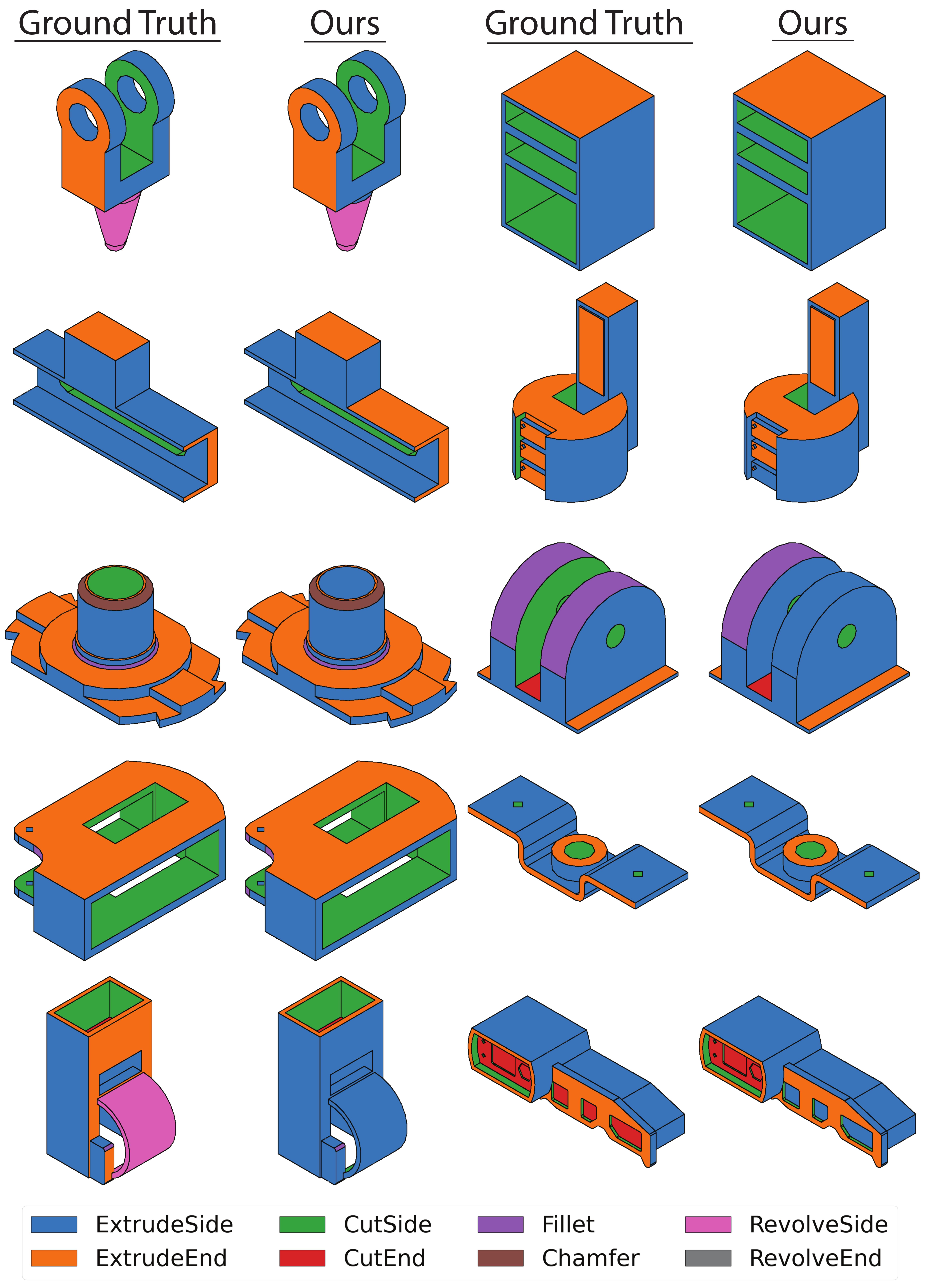}
    \caption{Face classification example results for the Fusion 360 Segmentation Task. The CAD operation which created each face is indicated by color. Left is our model's prediction, right is the ground truth.}
    \label{fig:f360examples}
\end{figure}

We pre-trained our geometric self-supervision network over this dataset, then our face-level prediction network using the face embeddings from the self-supervision. Some classification results are shown in Figure~\ref{fig:f360examples}. Our method is able to achieve 65\% accuracy after seeing just 10 training examples, and is 96\% accurate when given all 23266 in the training set.

We compare our method against two network architectures for B-Rep learning; BRepNet~\cite{lambourne_brepnet_2021} and UV-Net~\cite{jayaraman2021uv}. BRepNet defines a convolution operator relative to co-edges in a B-Rep structure and uses parametric face and edge types, as well as concavity and edge length features. UV-Net is a message passing network between adjacent faces, and uses grid sampled points of faces and edges as its features. We trained and tested each network on random subsets of the training data ranging from 10 examples to 23266. We repeated this 10 times at each training set size (all three methods saw the same training sets). 

\begin{figure}
    \centering
    \includegraphics[width=\linewidth,keepaspectratio]{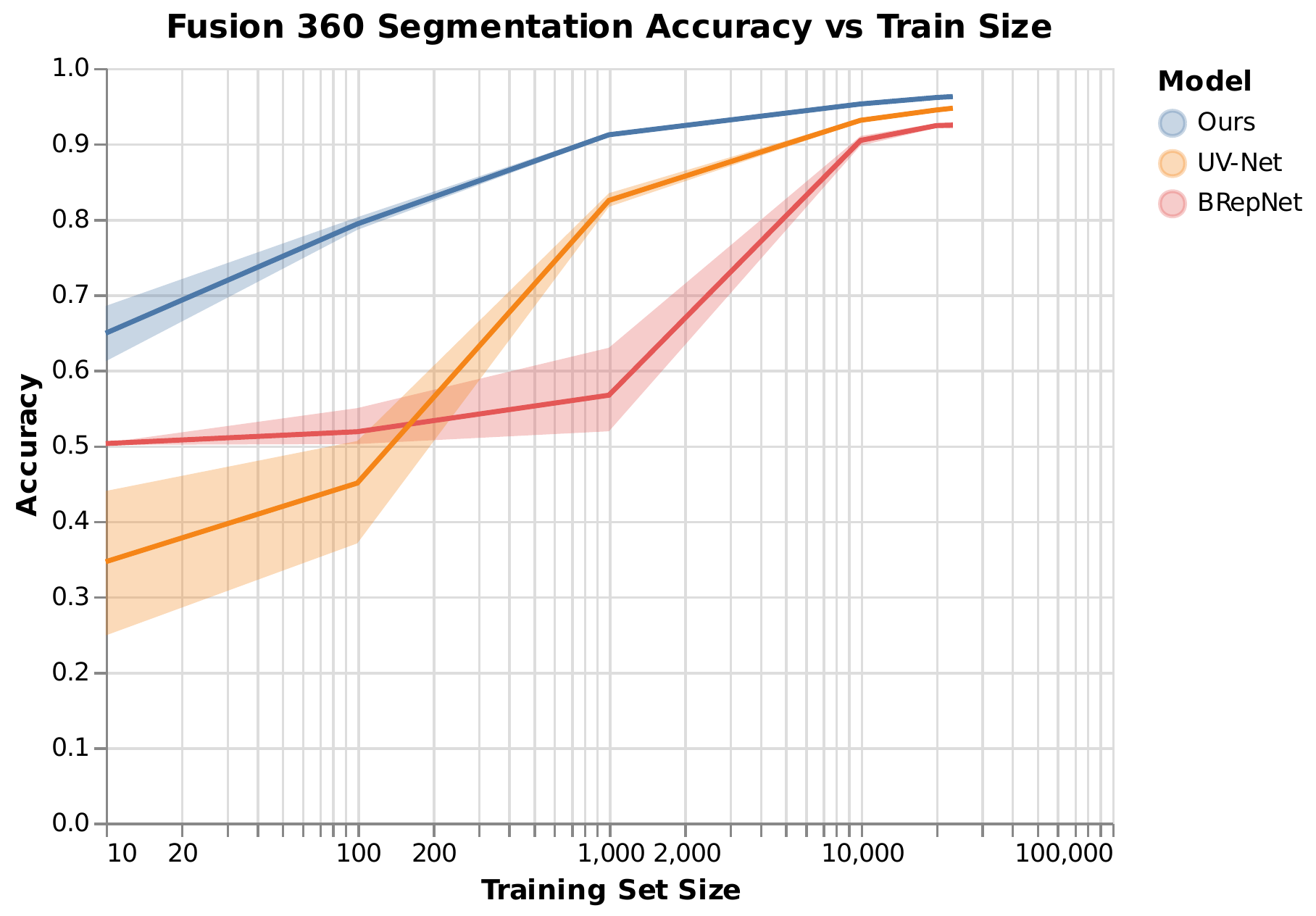}
    \caption{Comparison of our method versus UV-Net and BRepNet for Construction-Based Segmentation on the Fusion 360 Gallery Segmentation dataset. The mean accuracy across 10 training runs is plotted with a bootstrapped 95\% confidence interval.}
    \label{fig:f360plot}
\end{figure}

\begin{figure*}
    \centering
    \includegraphics[width= \linewidth,height=0.8\textheight,keepaspectratio]{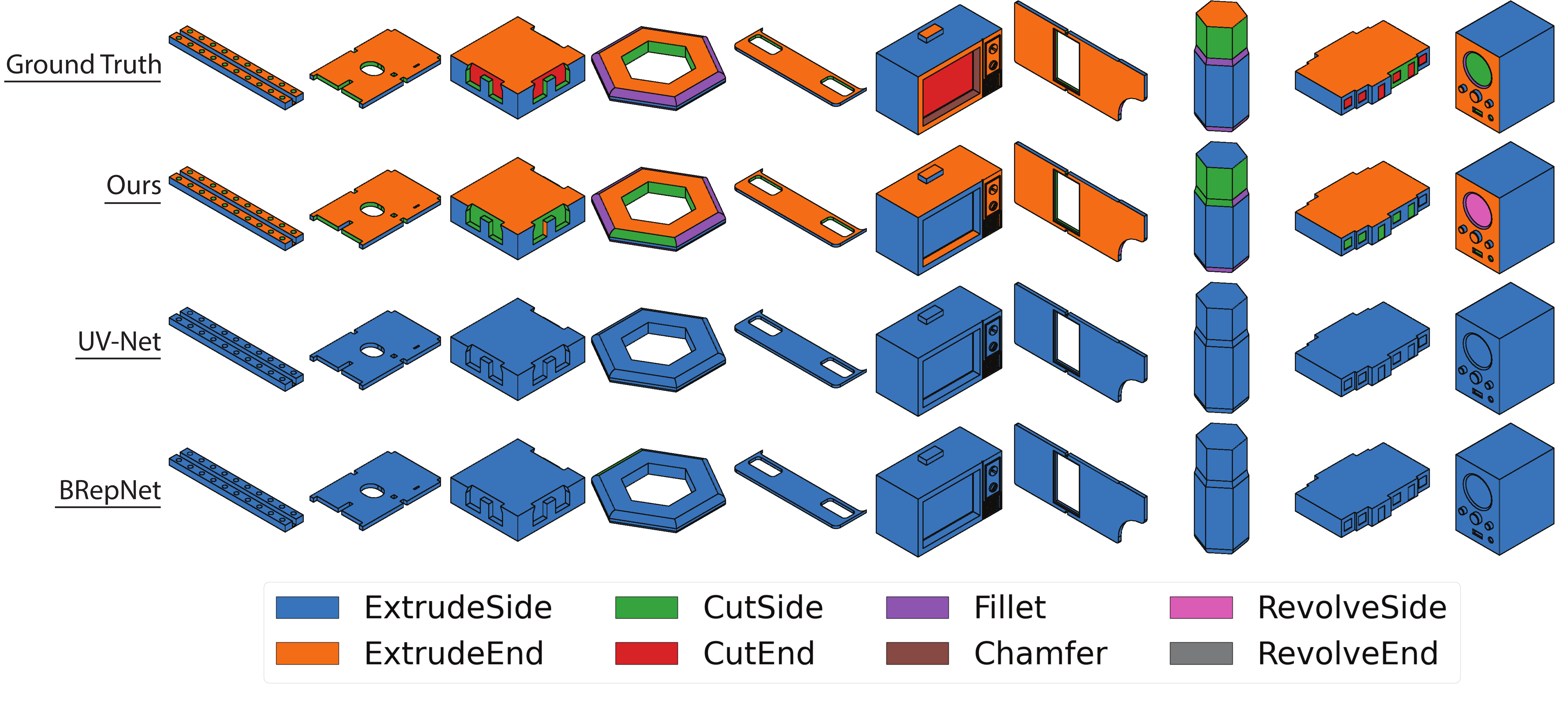}
    \caption{Few shot classification results from our method versus baseline models on the Fusion 360 gallery task, trained over just 100 models. Face color indicates the modeling operation used to create a face. Our model performs significantly better in the low data regime; it can differentiate operations, whereas the baselines are largely guessing the most common operation: ExtrudeSide.}
    \label{fig:classificationcomp}
\end{figure*}

Figure~\ref{fig:f360plot} plots the average accuracy at each train size across these runs with a bootstrapped 95\% confidence interval. Our method outperforms the baselines at all dataset sizes, and does so by a significant margin in the few-shot regime. It also has a smaller confidence interval, indicating that pre-training on our rasterization task makes classifications more robust to choice of training example\footnote{BRepNet's very small confidence interval at small training set sizes is an artifact of it predominantly or entirely predicting one class.}. In addition to these quantitative results, we show classification comparisons at the 100 example level in Figure~\ref{fig:classificationcomp}. Our method is able to start generalizing with 10s to 100s of examples, whereas the baselines fail to generalize and mostly learn the most frequent label at this data scale.

\begin{figure}
    \centering
    \includegraphics[width= \linewidth,height=0.8\textheight,keepaspectratio]{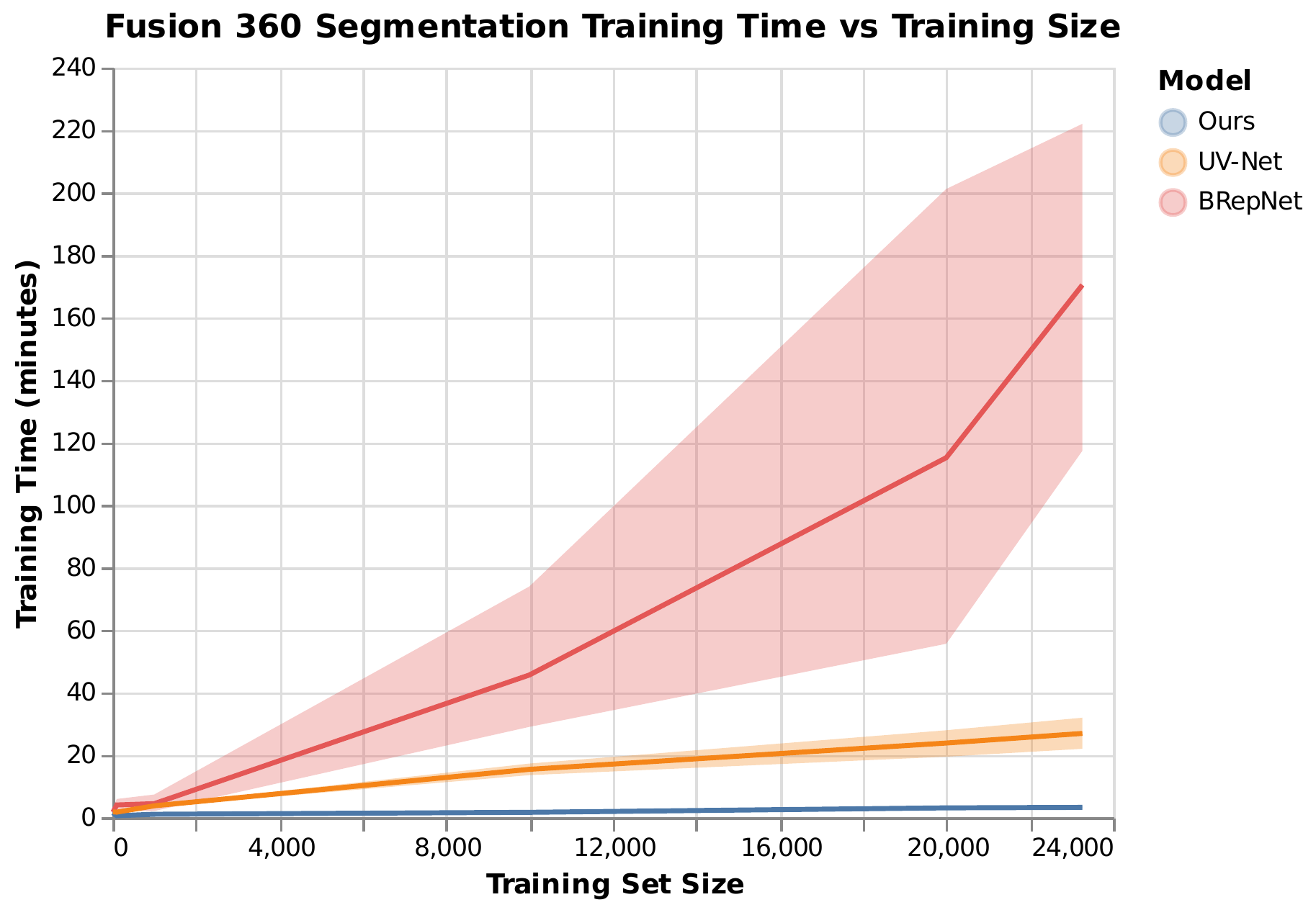}
    \caption{Training times on the Fusion 360 Segmentation task. All models were trained on an NVIDIA RTX 2080 Ti, and reported time is the time to minimum validation loss. Face codes pre-trained using a truncated SB-GCN.}
    \label{fig:timefig}
\end{figure}

We further note that training our method is an order of magnitude faster than either baseline (See Figure~\ref{fig:timefig}). This raises the possibility of segmentation tasks trained on-the-fly to enable predictors to be trained and deployed in the modeling process.

\subsection{Few-Shot Manufacturing-Driven Segmentation}
The second segmentation problem we considered is face segmentation that classifies how each face is manufactured. We evaluated this on the MFCAD dataset~\cite{cao2020graph}. This is a synthetic dataset of 15488 CAD models where each face is labeled with one of 16 types of machining feature (chamfer, through hole, blind hole, etc.) The parts in this dataset are generated by applying machining operations to square stock, and so consist entirely of planar faces and straight line edges. As compared to the Fusion 360 Gallery dataset, this task has twice as many classes to choose from, and will rely more on neighborhood and boundary information since all faces are planar. It also shows the ability of our approach to generalize pre-trained features to out-of-distribution data, since we use face codes pre-trained on Fusion 360 Gallery models (in this and all examples), which is more naturalistic in comparison.

\begin{figure*}
    \centering
    \includegraphics[width=\textwidth]{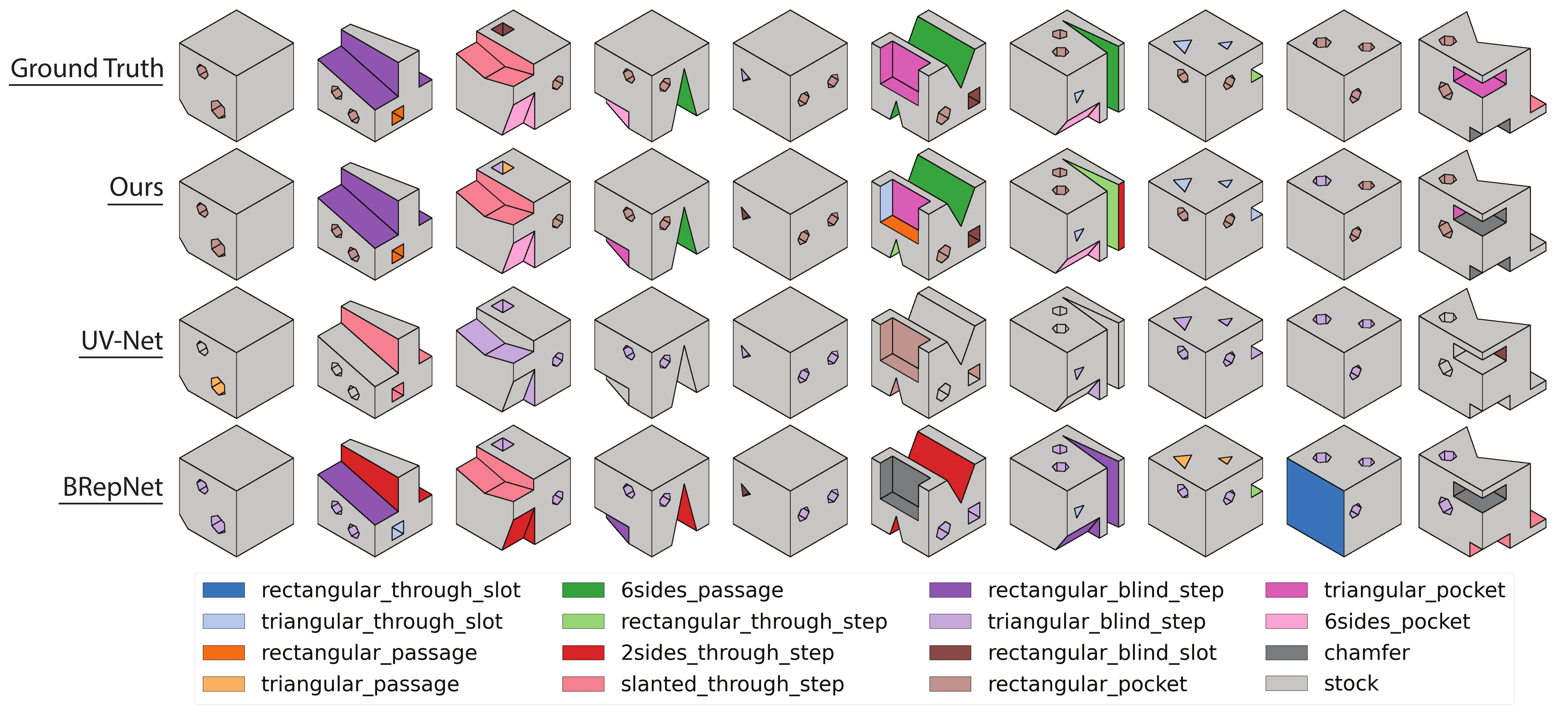}
    \caption{MFCAD few shot segmentation comparison at 100 train samples. Top row is the ground truth labeling, followed by Our, UV-Net, and BRepNet predictions.}
    \label{fig:mfcadtrainsizegallery}
\end{figure*}

\begin{figure}
    \centering
    \includegraphics[width=\linewidth,keepaspectratio]{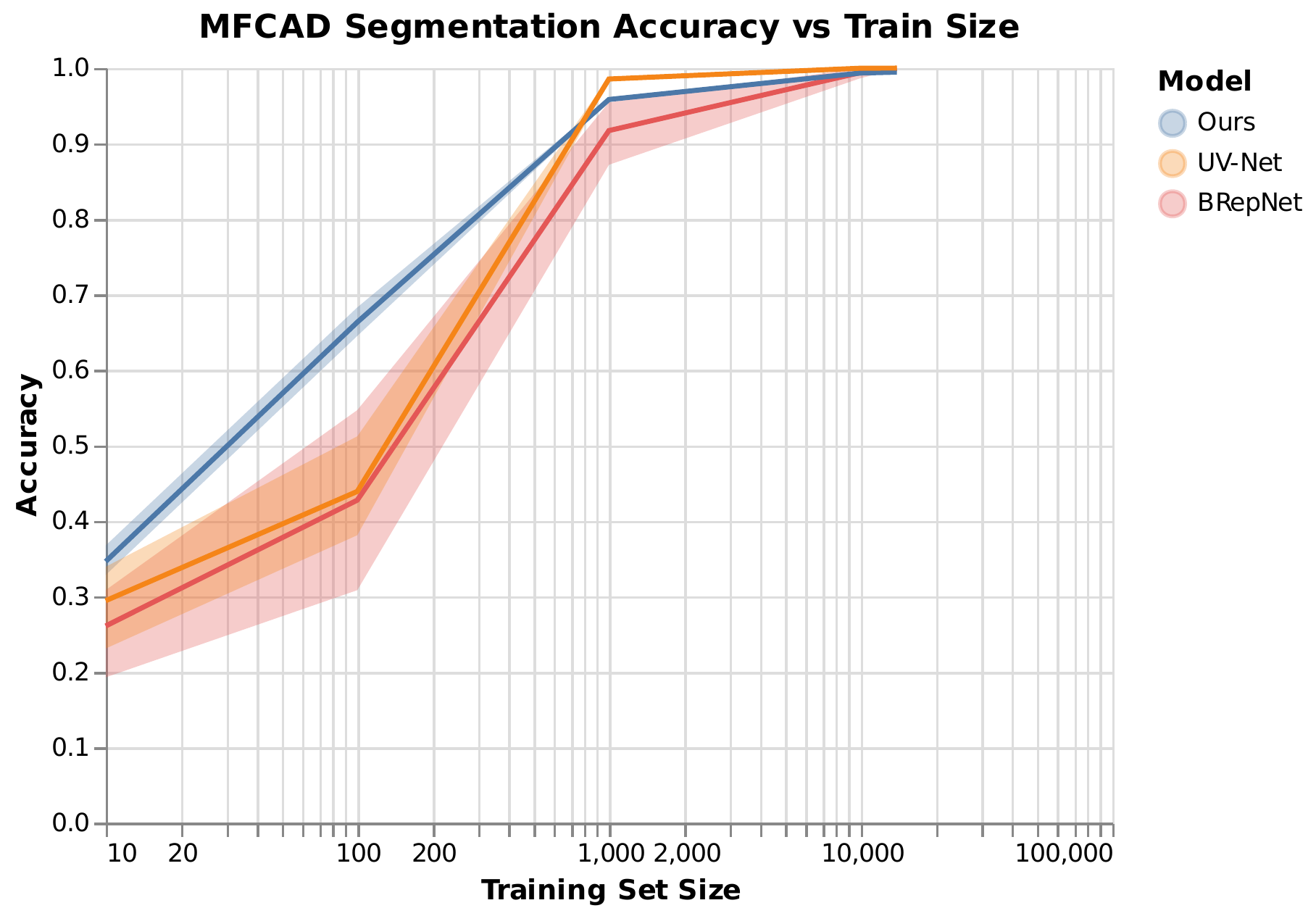}
    \caption{Comparison of our method versus UV-Net and BRepNet for Manufacturing-Based Segmentation on the MFCAD dataset. The mean accuracy across 10 training runs is plotted with a bootstrapped 95\% confidence interval.}
    \label{fig:mfCADplot}
\end{figure}

We compare this task to the same two baselines, illustrated qualitatively in Figure~\ref{fig:mfcadtrainsizegallery} and quantitatively in Figure~\ref{fig:mfCADplot}. We had to remove curvature features from BRepNet since they are universally zero on this dataset, and BRepNet requires all input features to have non-zero standard deviation. As before, our method outperforms the baselines at few-shot learning, and achieves comparable accuracy at large data sizes (all three methods are essentially perfect given sufficient training data). The stability of our prediction accuracy as measured by confidence interval is significantly better than both baselines.

\subsection{Few-Shot Part Classification}
In addition to segmentation, we also applied our method to part classification. For this, we used the FabWave~\cite{fabwave2019} dataset, a hand-labeled subset of GrabCAD~\cite{grabcad} which has been categorized into classes of mechanical parts (gears, brackets, washers, etc.) Many parts from within a category are parametric variations of each other. This task is important for understanding the function of mechanical parts in assemblies. 

\begin{figure}
    \centering
    \includegraphics[width=\linewidth,keepaspectratio]{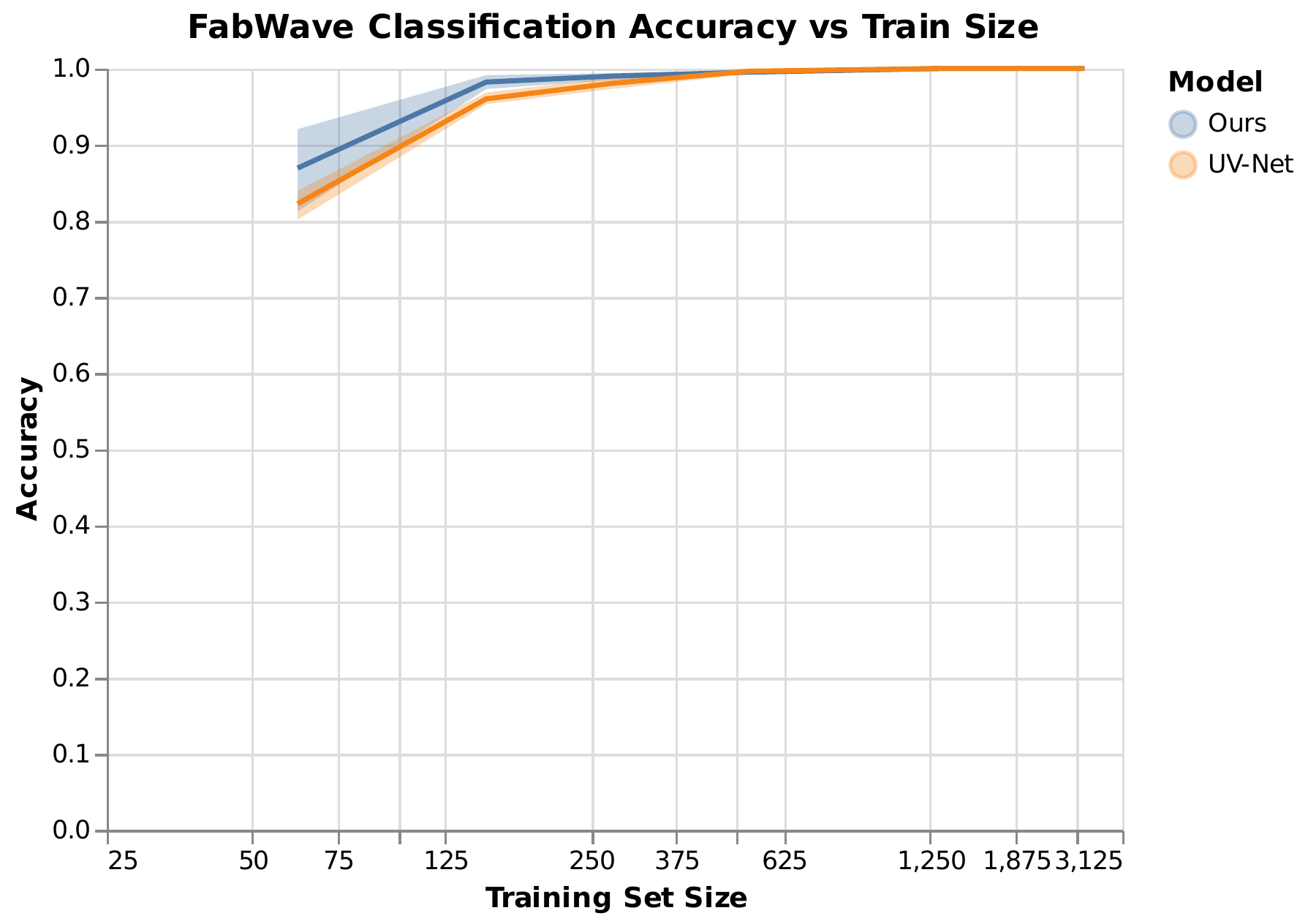}
    \caption{Comparison of our method versus UV-Net for Part Classification on the FabWave Dataset. The mean accuracy across 10 training runs is plotted with a bootstrapped 95\% confidence interval.}
    \label{fig:fabwaveplot}
\end{figure}

As a baseline comparison we only compare against UV-Net since BRepNet does not support classification. Again using face codes pre-trained on Fusion 360 Gallery B-Reps, we train our B-Rep Classification Network over the 26 classes which have at least 3 examples compatible with our network (for train, test, and validation), and use stratified sampling to create training subsets which contain at least one part from each category in each split. Since UV-Net cannot be run on B-Reps that do not contain edges, we restricted our test set to only models with edges, as well as removing 2 classes which UV-Net could not distinguish since they differ only in part orientation (our technique is able to distinguish these classes, so keeping them in the evaluation would both improve our accuracy and decrease UV-Net's). Figure~\ref{fig:fabwaveplot} shows the results. Our method outperforms UV-Net across training sizes and even achieves 100\% accuracy at higher training set sizes.

\subsection{Rasterization Evaluation}

\begin{figure*}
    \centering
    \includegraphics[width=\textwidth,keepaspectratio]{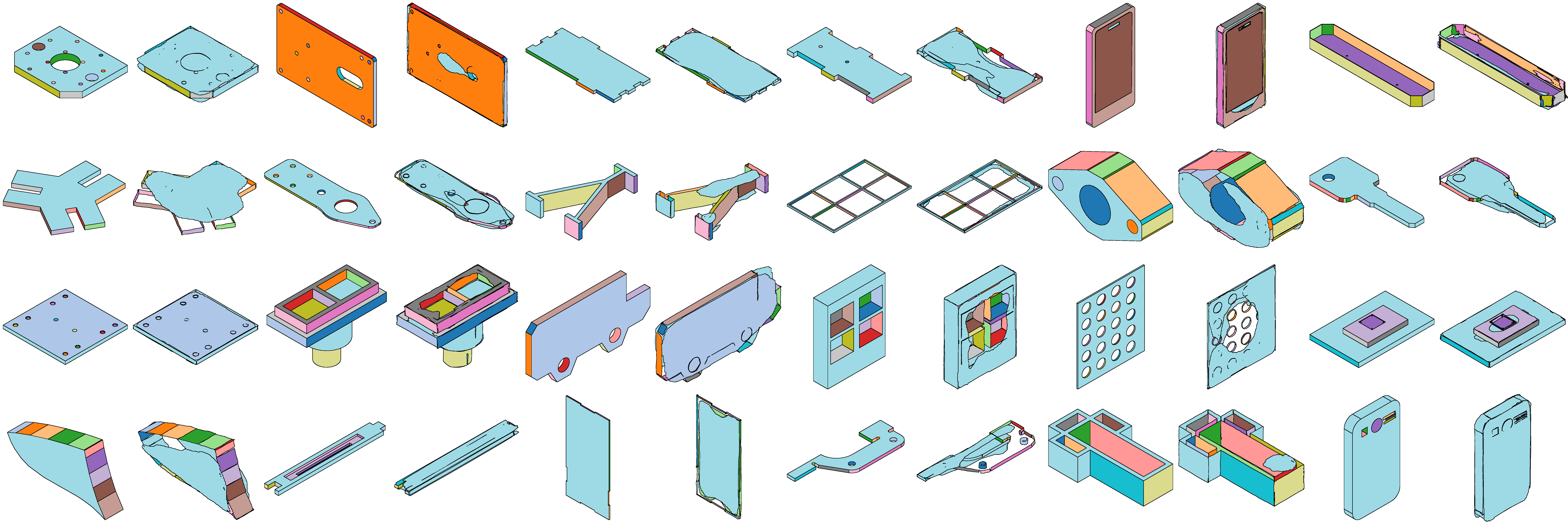}
    \caption{Shape reconstructions (right) from face embeddings on the Fusion 360 Segmentation test set, compared to ground truth (left). Each B-Rep face is given a unique color, which is consistent between ground truth and reconstruction. Reconstructions were created by sampling a 100x100 (u,v) grid in the range [-0.1,1.1] for each face to create a mesh for the supporting surface, then removing mesh vertices outside the predicted clipping plane SDF. Supporting surface reconstructions are highly accurate, as are the clipping masks for typical shapes with a single boundary. Complex interior and exterior boundaries of clipping planes are sometimes predicted inaccurately.}
    \label{fig:reconstructions}
\end{figure*}

In Figure~\ref{fig:reconstructions} we present a collection of part renderings to illustrate qualitatively our reconstruction and classification results. It shows a gallery of rasterization results on unseen test examples from the Fusion 360 Gallery segmentation dataset. We see that the explicit prediction of support surface coordinates works very well. While the implicit prediction of clipping planes works well in many cases, it sometimes misses inner loops and complex boundary geometry.

\subsection{Ablations}

\paragraph{Self-Supervision Ablations}
In addition to the network described in Section~\ref{sec:selfsup}, we also tried using a truncated SB-GCN (only using its upwards pass) as the encoder network, using the same shape parameter input features. We evaluate both explicit surface and SDF accuracy in Table~\ref{tab:selfsupablations}, and see that our architecture significantly outperforms SB-GCN in both measures.

\begin{table}[h]
	\centering
	\small
	\begin{tabular}{c c c}%
		Model & XYZ Error & SDF Error\\
		\toprule
		Ours & $\mathbf{0.0256}$ & $\mathbf{0.0147}$ \\
		SB-GCN & 0.035 & 0.0214 \\
		\bottomrule
	\end{tabular}
	\caption{Self-Supervision ablations. XYZ Error is the average pointwise distance between predicted and sampled surface position. SDF Error is the average absolute difference between predicted and actual SDF value.}
	\label{tab:selfsupablations}
\end{table}

\paragraph{Segmentation Ablations}
We tried 3 types of face-level prediction network using our self-supervised face embeddings to determine which was best. The first two try to directly classify faces from the embedding directly, one using a Linear Support Vector Machine (SVM), and the other using a Multi-Layer perceptron. These test linear and non-linear decision boundaries based on face-codes alone. The third is the message passing scheme described in Section~\ref{sec:fewshot}, which tests if neighborhood context is necessary.

To evaluate how well each method worked across labeled dataset sizes, we trained each model repeatedly on all supported tasks at a variety of training set sizes using a similar scheme to our baseline comparisons. Table~\ref{tab:segablations} records the average face segmentation accuracy across dataset sizes for both segmentation tasks. Results were consistent across datasets and training sizes; a non-linear decision boundary is more accurate than a linear one, and neighborhood information improves accuracy, leading us to choose the message passing network as our method.

\begin{table*}[t]
    \centering
    \small
    \begin{tabular}{@{}l l l c c c c c c @{}}%
\toprule%
\multicolumn{2}{c}{Task / Model}&\multicolumn{7}{c}{Training Set Size}\\%
\midrule%
\multicolumn{2}{l}{Fusion 360 Segmentation}&Accuracy@:&10&100&1000&10000&20000&23266\\%
\cmidrule(r){1%
-%
2}%
\cmidrule(l){3%
-%
9}%
&Self{-}Supervision + SVM&&$0.50$&$0.66$&$0.75$&$0.76$&$0.77$&$0.77$\\%
&Self{-}Supervision + MLP&&$0.55$&$0.67$&$0.82$&$0.90$&$0.91$&$0.91$\\%
&Self{-}Supervision + MP&&$\mathbf{0.56}$&$\mathbf{0.68}$&$\mathbf{0.83}$&$\mathbf{0.92}$&$\mathbf{0.94}$&$\mathbf{0.94}$\\%
\multicolumn{2}{l}{MFCAD}&Accuracy@:&10&100&1000&10000&13940&{-}{-}\\%
\cmidrule(r){1%
-%
2}%
\cmidrule(l){3%
-%
9}%
&Self{-}Supervision + SVM&&$0.15$&$0.50$&$0.56$&$0.59$&$0.58$&\\%
&Self{-}Supervision + MLP&&$0.35$&$0.55$&$0.83$&$0.90$&$0.91$&\\%
&Self{-}Supervision + MP&&$\mathbf{0.38}$&$\mathbf{0.65}$&$\mathbf{0.95}$&$\mathbf{0.99}$&$\mathbf{0.99}$&\\\bottomrule%
\end{tabular}
    \caption{Segmentation ablations. Reported face classification accuracies are the mean of 10 runs at each data set size with the train set subset at different random seeds (each model sees the same 10 random subsets). Model selected by best validation loss on a random 20\% validation split, except for the SVM models. Bold indicates indicates the best accuracy at each train size for each task. Self-Supervision codes pre-trained using a truncated SB-GCN.}
    \label{tab:segablations}
\end{table*}

\paragraph{Classification Ablations}
We also tested adding additional message passing layers prior to pooling for the classification network. As Table~\ref{tab:classablations} shows, this additional complexity did not yield real improvement on 

\begin{table*}[t]
    \centering
    \small
    \begin{tabular}{@{}l l l c c c c c c c @{}}%
\toprule%
\multicolumn{2}{c}{Task / Model}&\multicolumn{8}{c}{Training Set Fraction}\\%
\midrule%
\multicolumn{2}{l}{FabWave}&Accuracy@:&1\%&5\%&10\%&20\%&50\%&75\%&100\%\\%
\cmidrule(r){1%
-%
2}%
\cmidrule(l){3%
-%
10}%
&Self{-}Supervision + Pooling&&$\mathbf{0.72}$&$\mathbf{0.93}$&$0.97$&$\mathbf{0.99}$&$\mathbf{0.999}$&$.999$&$\mathbf{1.00}$\\
&Self{-}Supervision + MP + Pooling&&$0.72$&$0.93$&$\mathbf{0.97}$&$0.99$&$0.999$&$\mathbf{1.00}$&$\mathbf{1.00}$\\\bottomrule%
\end{tabular}
    \caption{Classification ablations. Reported accuracies are mean of 10 runs, similar to Table~\ref{tab:segablations}. Adding message passing prior to pooling does not confer an advantage, so we do not use it in our reported results. Self-Supervision codes were pre-trained using a truncated SB-GCN.}
    \label{tab:classablations}
\end{table*}

\subsection{Limitations and Future Work}
Our work has three main limitations. The first is the restriction of B-Reps to those with fixed-size parameter sets for all attached geometry. This is a limitation of the choice of encoder network, and could be alleviated by using a sequence or tree encoder to compute fixed-size embeddings for the generic functional geometric expressions of each B-Rep topology.

The second limitation is that we only self-supervise on local information. This means that we rely on additional message passing layers in our task specific decoders to gather neighborhood features.

Finally, we only create encodings for faces, which limits the kinds of tasks we can learn. Adding a second decoder and loss term to rasterize edges could extend this work to edge-based tasks, however there are not currently edge-specific tasks in the literature to compare against, and we encountered frequently CAD kernel errors when sampling edges, so we did not take this step.

\begin{figure}
    \centering
    \includegraphics[width=0.4\textwidth]{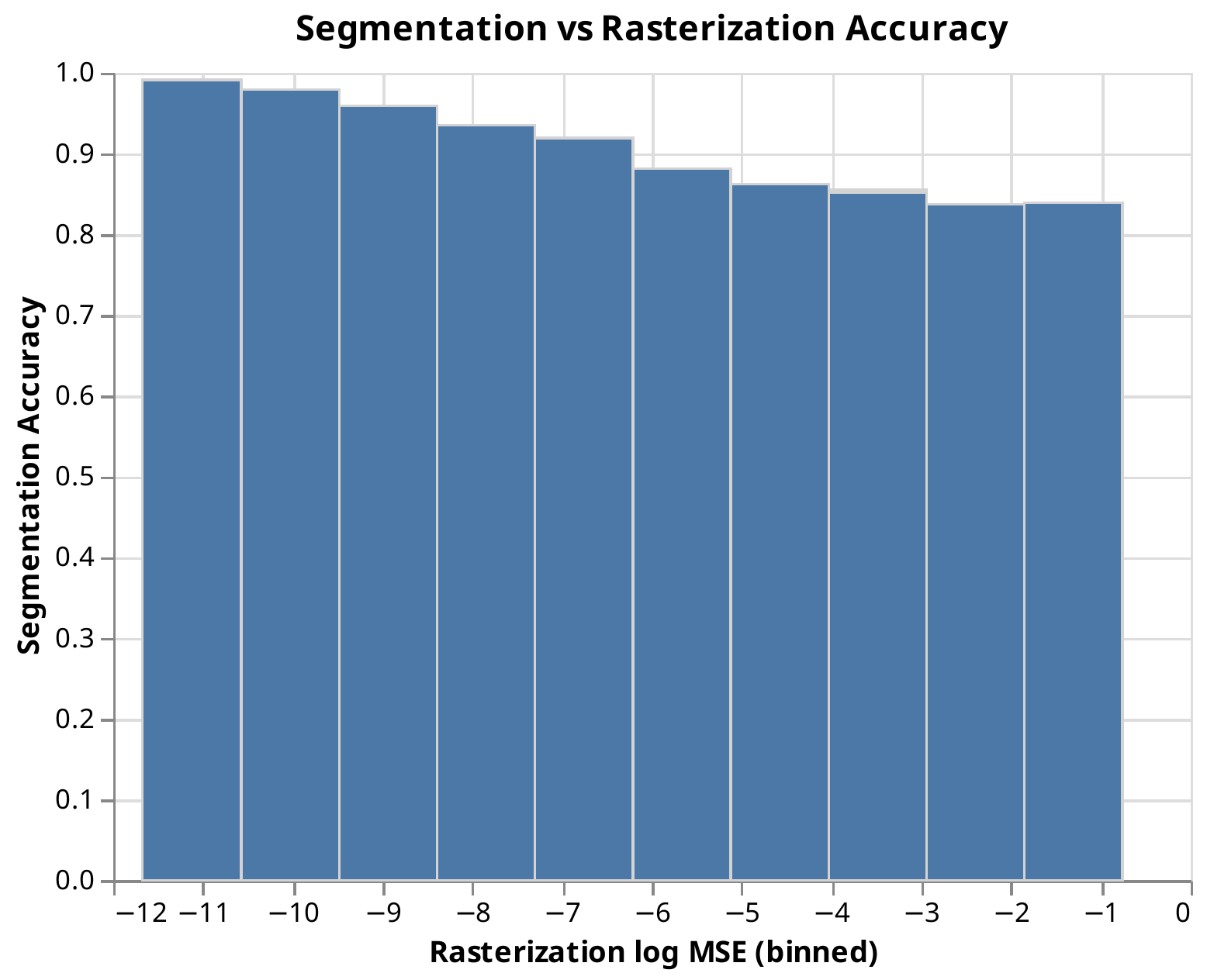}
    \caption{Segmentation accuracy compared to rasterization accuracy on the Fusion 360 segmentation task. X-axis is the log of the per-face MSE of our rasterization, binned into 10 groups, and the Y-axis is the fraction of faces segmented accurately within each bin. Data is aggregated across all segmentation training set sizes and seeds (the trend is similar across all sizes, with lower accuracies at lower training set sizes).}
    \label{fig:classifyvsreconstruction}
\end{figure}

Improving self-supervision accuracy will be a fruitful direction for future work. Figure~\ref{fig:classifyvsreconstruction} shows that the accuracy of downstream learning tasks is correlated with the accuracy of our model's rasterization. Thus improvements in rasterization performance should yield improvements in classification performance in both few-shot and large data regimes.

\begin{figure}
	\centering
	\includegraphics[width=0.4\textwidth]{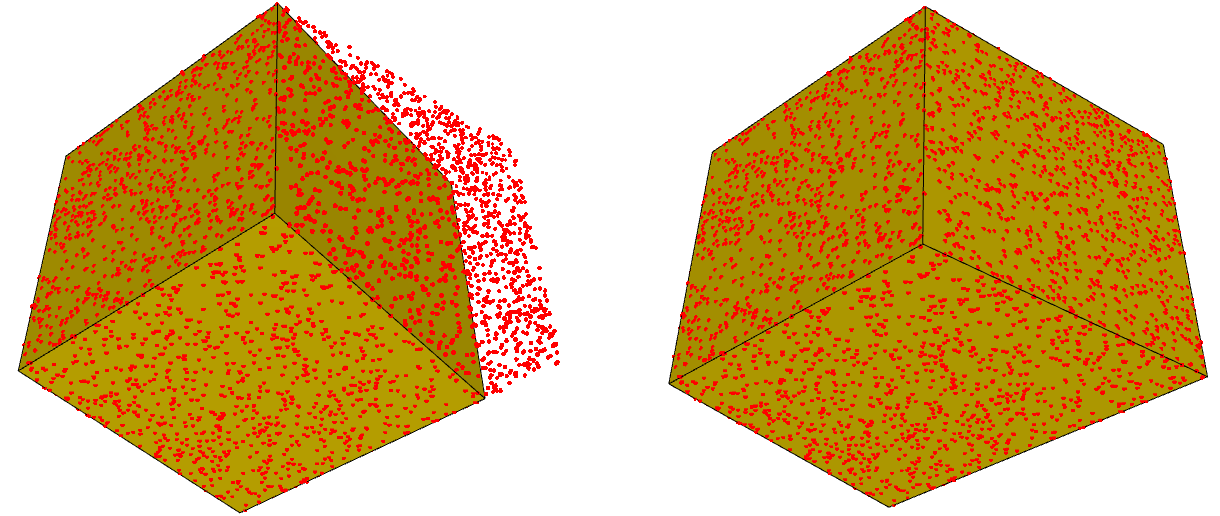}
	\caption{Shape matching with differentiable rasterization. The input cube (left) was optimized using SGD over its B-Rep shape parameters (right) to match a target point cloud (red).}
	\label{fig:gradientopt}
\end{figure}

Improving rasterization performance could also unlock future applications in reverse engineering. Since operating our self-supervision network as a rasterizer creates, in effect, a differentiable CAD renderer, we can use it for gradient-based optimization of B-Rep \emph{shape parameters}. To demonstrate the potential of such an application, we prototyped a shape matching application, where an input B-Rep is optimized via stochastic gradient descent to match a target point cloud. Figure~\ref{fig:gradientopt} shows the results of using this optimization to angle the face of a cube. We find that this optimization technique struggles on more complex shapes, which may overcome by improved rasterization performance.

\section{Conclusion}

In this work we propose to learn a spatial embedding of B-reps and apply it to few-short learning on supervised tasks. We validate that this approach is effective compared to prior work on supervised learning. Our results show comparable results on large training sets and significantly better performance on smaller sets.  By experimenting over three different tasks and data-sets, we posit that this method will be widely applicable on a plethora of CAD applications. Being faster to train can enable many applications in this domains; particularly, user-guided annotation for customized predictions. Finally, our method enables a fully differential embedding of B-rep geometry, compared to prior work that required non-differentiable CAD kernels, paving the way to exciting future work on CAD optimization and reverse engineering.

{\small
\bibliographystyle{ieee_fullname}
\bibliography{cad.bib}

\begin{thebibliography}{10}\itemsep=-1pt

\bibitem{grabcad}
Grabcad.
\newblock \url{https://grabcad.com/}.
\newblock Accessed: 2022-05-19.

\bibitem{parasolidCAD}
Siemens. parasolid cad kernel.
\newblock
  \url{https://www.plm.automation.siemens.com/global/en/products/plm-components/parasolid.html}.
\newblock Accessed: 2022-05-19.

\bibitem{fabwave2019}
{\em {Development of a Pilot Manufacturing Cyberinfrastructure With an
  Information Rich Mechanical CAD 3D Model Repository}}, volume Volume 1:
  Additive Manufacturing; Manufacturing Equipment and Systems; Bio and
  Sustainable Manufacturing of {\em International Manufacturing Science and
  Engineering Conference}, 06 2019.
\newblock V001T02A035.

\bibitem{achlioptas2017latent_pc}
Panos Achlioptas, Olga Diamanti, Ioannis Mitliagkas, and Leonidas~J Guibas.
\newblock Learning representations and generative models for 3d point clouds.
\newblock {\em arXiv preprint arXiv:1707.02392}, 2017.

\bibitem{DBLP:journals/corr/BrockLRW16}
Andr{\'{e}} Brock, Theodore Lim, James~M. Ritchie, and Nick Weston.
\newblock Generative and discriminative voxel modeling with convolutional
  neural networks.
\newblock {\em CoRR}, abs/1608.04236, 2016.

\bibitem{cao2020graph}
Weijuan Cao, Trevor Robinson, Yang Hua, Flavien Boussuge, Andrew~R. Colligan,
  and Wanbin Pan.
\newblock Graph representation of 3d cad models for machining feature
  recognition with deep learning.
\newblock volume Volume 11A: 46th Design Automation Conference (DAC) of {\em
  International Design Engineering Technical Conferences and Computers and
  Information in Engineering Conference}, 08 2020.

\bibitem{chen2019implicitfields}
Zhiqin Chen and Hao Zhang.
\newblock Learning implicit fields for generative shape modeling.
\newblock {\em IEEE Computer Vision and Pattern Recognition (CVPR)}, 2019.

\bibitem{doersch2015unsupervised}
Carl Doersch, Abhinav Gupta, and Alexei~A Efros.
\newblock Unsupervised visual representation learning by context prediction.
\newblock In {\em Proceedings of the IEEE International Conference on Computer
  Vision}, pages 1422--1430, 2015.

\bibitem{fan2017}
Haoqiang Fan, Hao Su, and Leonidas Guibas.
\newblock A point set generation network for 3d object reconstruction from a
  single image.
\newblock {\em CVPR}, 2017.

\bibitem{ganin2021computer}
Yaroslav Ganin, Sergey Bartunov, Yujia Li, Ethan Keller, and Stefano Saliceti.
\newblock Computer-aided design as language.
\newblock {\em Advances in Neural Information Processing Systems}, 34, 2021.

\bibitem{gidaris2018unsupervised}
Spyros Gidaris, Praveer Singh, and Nikos Komodakis.
\newblock Unsupervised representation learning by predicting image rotations.
\newblock In {\em ICLR 2018}, 2018.

\bibitem{groueix2018atlasnet}
Thibault Groueix, Matthew Fisher, Vladimir~G Kim, Bryan~C Russell, and Mathieu
  Aubry.
\newblock Atlasnet: A papier-mache approach to learning 3d surface generation.
\newblock {\em arXiv preprint arXiv:1802.05384}, 2018.

\bibitem{Guo2022Complexgen}
Hao-Xiang Guo, Shilin Liu, Hao Pan, Liu Yang, Xin Tong, and Baining Guo.
\newblock Complexgen: Cad reconstruction by b-rep chain complex generation.
\newblock {\em ACM Transactions on Graphics (TOG)}, 39(4):106:1--106:14, 2022.

\bibitem{jayaraman2022solidgen}
Pradeep~Kumar Jayaraman, Joseph~G Lambourne, Nishkrit Desai, Karl~DD Willis,
  Aditya Sanghi, and Nigel~JW Morris.
\newblock Solidgen: An autoregressive model for direct b-rep synthesis.
\newblock {\em arXiv preprint arXiv:2203.13944}, 2022.

\bibitem{jayaraman2021uv}
Pradeep~Kumar Jayaraman, Aditya Sanghi, Joseph~G Lambourne, Karl~DD Willis,
  Thomas Davies, Hooman Shayani, and Nigel Morris.
\newblock Uv-net: Learning from boundary representations.
\newblock In {\em Proceedings of the IEEE/CVF Conference on Computer Vision and
  Pattern Recognition}, pages 11703--11712, 2021.

\bibitem{jones:2021:automate}
Benjamin Jones, Dalton Hildreth, Duowen Chen, Ilya Baran, Vladimir~G. Kim, and
  Adriana Schulz.
\newblock Automate: A dataset and learning approach for automatic mating of cad
  assemblies.
\newblock {\em ACM Transactions on Graphics}, 40(6), dec 2021.

\bibitem{kingma2014adam}
Diederik~P Kingma and Jimmy Ba.
\newblock Adam: A method for stochastic optimization.
\newblock {\em arXiv preprint arXiv:1412.6980}, 2014.

\bibitem{kingma2014semi}
Durk~P Kingma, Shakir Mohamed, Danilo~Jimenez Rezende, and Max Welling.
\newblock Semi-supervised learning with deep generative models.
\newblock In {\em Advances in neural information processing systems}, pages
  3581--3589, 2014.

\bibitem{kingma2013auto}
Diederik~P Kingma and Max Welling.
\newblock Auto-encoding variational bayes.
\newblock {\em arXiv preprint arXiv:1312.6114}, 2013.

\bibitem{koch_abc_2019}
Sebastian Koch, Albert Matveev, Zhongshi Jiang, Francis Williams, Alexey
  Artemov, Evgeny Burnaev, Marc Alexa, Denis Zorin, and Daniele Panozzo.
\newblock {ABC}: {A} {Big} {CAD} {Model} {Dataset} for {Geometric} {Deep}
  {Learning}.
\newblock In {\em 2019 {IEEE}/{CVF} {Conference} on {Computer} {Vision} and
  {Pattern} {Recognition} ({CVPR})}, pages 9593--9603, Long Beach, CA, USA,
  June 2019. IEEE.

\bibitem{lambourne_brepnet_2021}
Joseph~G. Lambourne, Karl D.~D. Willis, Pradeep~Kumar Jayaraman, Aditya Sanghi,
  Peter Meltzer, and Hooman Shayani.
\newblock {BRepNet}: {A} topological message passing system for solid models.
\newblock {\em arXiv:2104.00706 [cs]}, Apr. 2021.
\newblock arXiv: 2104.00706.

\bibitem{larsson2016learning}
Gustav Larsson, Michael Maire, and Gregory Shakhnarovich.
\newblock Learning representations for automatic colorization.
\newblock In {\em European Conference on Computer Vision}, pages 577--593.
  Springer, 2016.

\bibitem{li2020sketch2cad}
Changjian Li, Hao Pan, Adrien Bousseau, and Niloy~J Mitra.
\newblock Sketch2cad: Sequential cad modeling by sketching in context.
\newblock {\em ACM Transactions on Graphics (TOG)}, 39(6):1--14, 2020.

\bibitem{liu2018voxelgan}
Jerry Liu, Fisher Yu, and Thomas Funkhouser.
\newblock Interactive 3d modeling with a generative adversarial network.
\newblock {\em International Conference on 3D Vision (3DV)}, 2017.

\bibitem{mescheder2019occupancy}
Lars Mescheder, Michael Oechsle, Michael Niemeyer, Sebastian Nowozin, and
  Andreas Geiger.
\newblock Occupancy networks: Learning 3d reconstruction in function space.
\newblock In {\em Proceedings of the IEEE Conference on Computer Vision and
  Pattern Recognition}, pages 4460--4470, 2019.

\bibitem{noroozi2016unsupervised}
Mehdi Noroozi and Paolo Favaro.
\newblock Unsupervised learning of visual representations by solving jigsaw
  puzzles.
\newblock In {\em European Conference on Computer Vision}, pages 69--84.
  Springer, 2016.

\bibitem{palmer2022deepcurrents}
David Palmer, Dmitriy Smirnov, Stephanie Wang, Albert Chern, and Justin
  Solomon.
\newblock {DeepCurrents}: Learning implicit representations of shapes with
  boundaries.
\newblock In {\em Proceedings of the IEEE/CVF Conference on Computer Vision and
  Pattern Recognition (CVPR)}, 2022.

\bibitem{para2021sketchgen}
Wamiq Para, Shariq Bhat, Paul Guerrero, Tom Kelly, Niloy Mitra, Leonidas~J
  Guibas, and Peter Wonka.
\newblock Sketchgen: Generating constrained cad sketches.
\newblock {\em Advances in Neural Information Processing Systems}, 34, 2021.

\bibitem{park2019deepsdf}
Jeong~Joon Park, Peter Florence, Julian Straub, Richard~A. Newcombe, and Steven
  Lovegrove.
\newblock Deepsdf: Learning continuous signed distance functions for shape
  representation.
\newblock {\em CVPR}, 2019.

\bibitem{pathak2016context}
Deepak Pathak, Philipp Krahenbuhl, Jeff Donahue, Trevor Darrell, and Alexei~A
  Efros.
\newblock Context encoders: Feature learning by inpainting.
\newblock In {\em Proceedings of the IEEE conference on computer vision and
  pattern recognition}, pages 2536--2544, 2016.

\bibitem{seff_sketchgraphs_2020}
Ari Seff, Yaniv Ovadia, Wenda Zhou, and Ryan~P. Adams.
\newblock {SketchGraphs}: {A} {Large}-{Scale} {Dataset} for {Modeling}
  {Relational} {Geometry} in {Computer}-{Aided} {Design}.
\newblock {\em arXiv:2007.08506 [cs, stat]}, July 2020.
\newblock arXiv: 2007.08506.

\bibitem{seff2021vitruvion}
Ari Seff, Wenda Zhou, Nick Richardson, and Ryan~P Adams.
\newblock Vitruvion: A generative model of parametric cad sketches.
\newblock {\em arXiv preprint arXiv:2109.14124}, 2021.

\bibitem{liu2018meshVAE}
Qingyang Tan, Lin Gao, Yu-Kun Lai, and Shihong Xia.
\newblock Variational autoencoders for deforming 3d mesh models.
\newblock In {\em Proceedings of the IEEE Conference on Computer Vision and
  Pattern Recognition}, 2018.

\bibitem{yaqing_fewshot}
Yaqing Wang, Quanming Yao, James~T. Kwok, and Lionel~M. Ni.
\newblock Generalizing from a few examples: A survey on few-shot learning.
\newblock {\em ACM Comput. Surv.}, 53(3), jun 2020.

\bibitem{willis2021joinable}
Karl~DD Willis, Pradeep~Kumar Jayaraman, Hang Chu, Yunsheng Tian, Yifei Li,
  Daniele Grandi, Aditya Sanghi, Linh Tran, Joseph~G Lambourne, Armando
  Solar-Lezama, et~al.
\newblock Joinable: Learning bottom-up assembly of parametric cad joints.
\newblock {\em arXiv preprint arXiv:2111.12772}, 2021.

\bibitem{willis_fusion_2020}
Karl D.~D. Willis, Yewen Pu, Jieliang Luo, Hang Chu, Tao Du, Joseph~G.
  Lambourne, Armando Solar-Lezama, and Wojciech Matusik.
\newblock Fusion 360 {Gallery}: {A} {Dataset} and {Environment} for
  {Programmatic} {CAD} {Reconstruction}.
\newblock {\em arXiv:2010.02392 [cs]}, Oct. 2020.
\newblock arXiv: 2010.02392.

\bibitem{wu2021deepcad}
Rundi Wu, Chang Xiao, and Changxi Zheng.
\newblock Deepcad: A deep generative network for computer-aided design models.
\newblock In {\em Proceedings of the IEEE/CVF International Conference on
  Computer Vision}, pages 6772--6782, 2021.

\bibitem{yang2018foldingnet}
Yaoqing Yang, Chen Feng, Yiru Shen, and Dong Tian.
\newblock Foldingnet: Point cloud auto-encoder via deep grid deformation.
\newblock In {\em Proceedings of the IEEE Conference on Computer Vision and
  Pattern Recognition (CVPR)}, volume~3, 2018.

\bibitem{you2020designspace}
Jiaxuan You, Rex Ying, and Jure Leskovec.
\newblock Design space for graph neural networks.
\newblock In {\em Proceedings of the 34th International Conference on Neural
  Information Processing Systems}, NIPS'20, Red Hook, NY, USA, 2020. Curran
  Associates Inc.

\bibitem{zhang2016colorful}
Richard Zhang, Phillip Isola, and Alexei~A Efros.
\newblock Colorful image colorization.
\newblock In {\em European conference on computer vision}, pages 649--666.
  Springer, 2016.

\end{thebibliography}
}

\end{document}